\documentclass[12pt]{article}
\usepackage{amssymb}                         % Дополнительные символы
\usepackage[T2A]{fontenc}
\usepackage{enumerate,amsmath,graphicx}
\usepackage{graphicx}

\begin{document}

\pagestyle{plain}
\title{Genetic Algorithm with Optimal Recombination for the Asymmetric Travelling Salesman Problem}

\author{Anton V. Eremeev, Yulia V. Kovalenko\thanks{This research is supported by the Russian Science
Foundation grant 15-11-10009.}}

\maketitle
\begin{center}
Sobolev Institute of Mathematics, \\
4, Akad. Koptyug avenue,\\
630090, Novosibirsk, Russia.\\
Email:~eremeev@ofim.oscsbras.ru,~julia.kovalenko.ya@yandex.ru
\end{center}

\begin{abstract}
We propose a new genetic algorithm with optimal recombination for
the asymmetric instances of travelling salesman problem. The
algorithm incorporates several new features that contribute to its
effectiveness: 1.~Optimal recombination problem is solved within
crossover operator. 2.~A new mutation operator performs a random
jump within 3-opt or 4-opt neighborhood. 3.~Greedy constructive
heuristic of W.~Zhang and 3-opt local search heuristic are used to
generate the initial population. A computational experiment on
TSPLIB instances shows that the proposed algorithm yields
competitive results to other well-known memetic algorithms for
asymmetric travelling salesman problem.
\\
%Some variations of the ORP formulation are discussed.

\end{abstract}

%-----------------------------------------------------------------
\section{Introduction} \label{sec:intro}
%-----------------------------------------------------------------
Travelling Salesman Problem~(TSP) is a
well-known NP-hard combinatorial optimization problem~\cite{GJ}.
Given a complete digraph~$G$  with the set of vertices
$V=\{v_1,\dots,v_n\}$, the set of arcs~$A=\{(v_i,v_j):\ v_i,v_j\in
V,\ i\ne j\}$ and arc weights (lengths)~$c_{ij}\ge 0$ of each
arc~$(v_i,v_j)\in A$, the TSP asks for a Hamiltonian circuit of
minimum length. If $c_{ij}\ne c_{ji}$ for at least one
pair~$(v_i,v_j)$ then the TSP is called the Asymmetric Travelling
Salesman Problem~(ATSP).
Numerous metaheuristics and heuristics have been proposed for the
TSP and the genetic algorithms~(GAs) are among them (see
e.g.~\cite{BFM2004,CS03,EK2014II,TWO14,YI}).

The performance of GAs depends significantly upon the choice of
the {\em crossover} operator, where the components of parent
solutions are combined to build the offspring. A supplementary
problem that emerges in some versions of crossover operator is
called {\em Optimal Recombination Problem}~(ORP). Given two
feasible parent solutions, ORP consists in finding the best
possible offspring in view of the basic principles of
crossover~\cite{R94}.
Experimental results~\cite{CS03,TWO14,YI} indicate that ORP may be used successfully in genetic algorithms.

In this paper, we propose a new GA using the ORP with
adjacency-based representation to solve the ATSP. Two simple
crossover-based GAs for~ATSP using ORPs were investigated
in~\cite{EK2014II} but no problem-specific local search procedures
or fine-tuning of parameters were used.
In comparison to the GAs from~\cite{EK2014II}, the GA proposed in
this paper uses a 3-opt local search heuristic and a
problem-specific heuristic of W.~Zhang~\cite{Zhang00} to generate
the initial population. In addition, this GA applies a new
mutation operator, which performs a random jump within 3-opt or
4-opt neighborhood. The current GA is based on the steady state
replacement~\cite{Reev1997}, while the GAs in~\cite{EK2014II} were
based on the elitist recombination (see e.g.~\cite{GoldTheir94}).
The experimental evaluation on instances from TSPLIB library shows
that the proposed GA  yields results competitive to those obtained
by some other well-known evolutionary algorithms for the ATSP.
%-----------------------------------------------------------------
\section{Genetic Algorithm} \label{sec:GA}
%-----------------------------------------------------------------
The genetic algorithm is a random search method
that models a process of evolution of a population of {\em
individuals}~\cite{Reev1997}. Each individual is a sample solution
to the optimization problem being solved.
The components of an individual are called {\em genes}.
Individuals of a new population are built by means of reproduction
operators (crossover and/or mutation). The crossover operator
produces the offspring from two parent individuals by combining
and exchanging their genes. The mutation adds small random changes
to an individual.

The formal scheme of the GA with steady state replacement is as follows:\\
\hspace{0cm}\vspace{-0.3cm}

{\bf Steady State Genetic Algorithm}

{Step~1}.~Construct the initial population and assign $t:=1$.

{Step~2}.~Repeat steps~2.1-2.4 until some stopping criterion is
satisfied:

\mbox{\hspace{1.5em}} { 2.1.}~Choose two parent individuals~${\bf
p}_1, {\bf p}_2$ from the population.

\mbox{\hspace{1.5em}} { 2.2.}~Apply mutation to ${\bf p}_1$ and
${\bf p}_2$ and obtain individuals~${\bf p}'_1, {\bf p}'_2$.

\mbox{\hspace{1.5em}} { 2.3.}~Create an offspring ${\bf p}'$,
applying a crossover to~${\bf p}'_1$ and ${\bf p}'_2$.

\mbox{\hspace{1.5em}} { 2.4.}~Choose a least fit individual in
population and replace it by~${\bf p}'$.

\mbox{\hspace{1.5em}} { 2.5.}~Set $t:=t+1$.

{ Step~3}.~The result is the best found individual w.r.t. objective function.\\
\hspace{0cm}\vspace{-0.3cm}

Our implementation of the GA is initiated by generating $N$
initial solutions, and the population size~$N$ remains constant
during the execution of the GA. Two individuals of the initial
population are constructed by means of the problem-specific
heuristic of W.~Zhang~\cite{Zhang00}. The heuristic first solves
the Assignment Problem, and then patches the cycles of the optimum
assignment together to form a feasible tour.
R.~Karp~\cite{Karp1979} proposed two variants of the patching. In
the first one, some cycle of maximum length is selected and the
remaining cycles are patched into it. In the second one, cycles
are patched one by one in a special sequence, starting with a
shortest cycle. All other $N-2$ individuals of the initial
population are generated using the {\em arbitrary insertion}
method~\cite{YI}, followed by a local search heuristic with a
3-opt neighborhood (see Subsection~\ref{subsec:LSH}).

Each parent on Step~2.1 is chosen by $s$-{\em tournament
selection}: sample randomly $s$~individuals from the current
population and select a fittest among them.

Operators of crossover and mutation are described in
Subsections~\ref{subsec:recombination} and~\ref{subsec:mutation}.
%-----------------------------------------------------------------
\subsection{Recombination Operators}\label{subsec:recombination}
%-----------------------------------------------------------------
Suppose that feasible solutions to the ATSP  are encoded as
vectors of adjacencies, where the immediate predecessor is
indicated for each vertex. Then the {\em optimal recombination
problem} with adjacency-based representation~\cite{EK2014IIY}
consists in finding a shortest travelling salesman's tour which
coincides with two given feasible parent solutions in arcs
belonging to both solutions and does not contain the arcs absent
in both solutions. These constraints are equivalent to a
requirement that the recombination should be respectful and gene
transmitting as defined in~\cite{R94}. The ORP with
adjacency-based representation for the  ATSP  is shown to be
NP-hard but it can be reduced to the TSP on graphs with bounded
vertex degrees~\cite{EK2014IIY}. The resulting TSP may be solved
in~${O(n 2^{\frac{n}{2}})}$ time by means of an adaptation of the
algorithm proposed by D.~Eppstein~\cite{Epp2007}.  A detailed
description of the reduction can be found in~\cite{EK2014IIY}. An
experimental evaluation of the ORP with adjacency-based
representation in a crossover-based GA was carried out
in~\cite{EK2014II}. The experiments showed that the CPU cost of
solving the ORPs in this GA is acceptable and decreases with
iterations count, due to decreasing population diversity. In what
follows, the optimized crossover operator, which solves the ORP
with adjacency-based representation will be called Optimized
Directed Edge Crossover~(ODEC). This operator may be considered as
a deterministic ``direct descendant'' of Directed Edge Crossover
(DEC)~\cite{S91}. Unlike DEC, Optimized Directed Edge Crossover
guarantees genes transmission.

An alternative way for solution encoding to the ATSP is the
position-based representation, where a feasible solution is
encoded as a sequence of the vertices of the TSP tour. The
computational experiment performed in~\cite{EK2014II} indicates
that the ORP for the adjacency-based representation has an
advantage over the ORP for the position-based representation on
ATSP instances from TSPLIB library. Therefore, in this paper we
consider only the adjacency-based representation.

Note that most of the known GAs for the TSP (see e.g.
\cite{BFM2004,Johnson97,TWO14}) apply a local search on GA
iterations. However an optimal recombination may be considered as
a best-improving move in a neighborhood defined by two parent
solutions. So we use a local search only at the initialization
stage.
%-----------------------------------------------------------------
\subsection{Local Search Heuristic} \label{subsec:LSH}
%-----------------------------------------------------------------
In general, $k$-opt  neighborhood for TSP is defined as the set of
tours that can be obtained from a given tour by replacing~$k$
arcs. Our Local Search Heuristic is a typical local search
heuristic that explores a subset of  $3$-opt
neighborhood.

\begin{figure}[!h]
\begin{center}
\includegraphics[width=8.27cm,height=4cm]{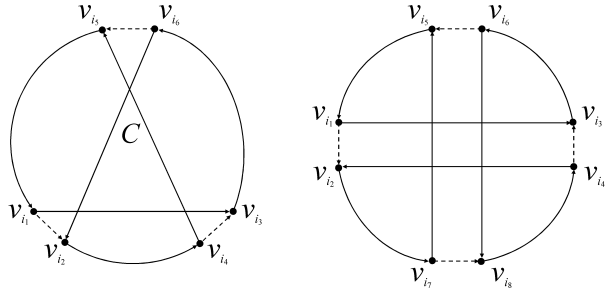}
\caption{3-change and quad change.} \label{fig:change}
\end{center}
\end{figure}

We try to improve the current tour by changing three of its arcs
(see Figure~\ref{fig:change}). To this end, we consider all
possible arcs of the current  tour as candidates for arc
$(v_{i_1},v_{i_2})$ to be deleted in the order of decreasing
length. Observe that, in our search, the possibilities for
choosing~$v_{i_3}$ (arc $(v_{i_1},v_{i_3})$ is added) may be
limited to those vertices that are closer to $v_{i_1}$ than
$v_{i_2}$. To use this property, for each vertex $v$ we store a
list of the remaining vertices in the order of increasing length
from~$v$. Considering candidates for~$v_{i_3}$, we start at the
beginning of $v_{i_1}$'s list and proceed down the list until a
vertex~$u$ with $c_{v_{i_1},u} \ge c_{v_{i_1},v_{i_2}}$ is
reached. Moreover, only the $\lceil0.2n\rceil$  nearest vertices
are stored in the sorted list of each vertex, which allows to
reduce the running time and the memory usage as observed
in~\cite{Johnson97}.
Finally, among all vertices belonging to the closed cycle~$C$
created by  $(v_{i_1},v_{i_3}),$  we choose a vertex~$v_{i_5}$
that would produce the most favorable 3-change. Local Search
Heuristic stops if no favorable 3-change is possible, otherwise it
proceeds to the next step with a new tour obtained.
\hspace{0cm}\vspace{-0.3cm}

\hspace{0cm}\vspace{-1.3cm}

In order to reduce the running time of  the presented local search
heuristic, we use the well-known  ``don’t look bits'' and ``first improving move'' strategies presented
in~\cite{Johnson97} for local search based on  $3$-opt neighborhood.

\subsection{Mutation Operators}\label{subsec:mutation}
The mutation is applied to each parent solution on Step~2.2  with
probability $p_{mut}$, which is a tunable parameter of the GA. We
implement two mutation operators that perform a random jump within
3-opt or 4-opt neighborhood. Each time one of the operators is
used for mutation with equal probability.

The first mutation operator makes a 3-change
(see Section~\ref{subsec:LSH}). First of all, an
arc~$(v_{i_1},v_{i_2})$ is chosen at random among all arcs of the
given tour. After that, an arc~$(v_{i_1},v_{i_3})$ is chosen using
the following idea from~\cite{KP1980}. For each possible arc
$(v_{i_1},u)$, we calculate $F(u)=c_{v,u}+|C(u)|\cdot c_{aver}$,
where $v$ is  the immediate predecessor of $u$ in the given tour,
$|C(u)|$ is the length of the cycle created by $u$ and $c_{aver}$
is the average weight of arcs in graph~$G$. Then an arc
$(v_{i_1},v_{i_3})$ is chosen uniformly at random among the top
$50\%$ of arcs $(v_{i_1},u)$ w.r.t. $F(u)$ value. The reason for
taking into account the value of~$|C(u)|$ when $(v_{i_1},v_{i_3})$
is being chosen is that the bigger the value of~$C(u)$ the more
options for $(v_{i_4},v_{i_5})$ will become available
subsequently. Finally, we choose the arc~$(v_{i_6},v_{i_5})$ among
the arcs of~$C(v_{i_3})$ so that the most favorable 3-change is
produced.

The second mutation operator is based on 4-opt neighborhood and
implements a {\em quad change}~\cite{KP1980} (see
Figure~\ref{fig:change}). Here two arcs $(v_{i_1},v_{i_2})$ and
$(v_{i_7},v_{i_8})$ are selected randomly and removed, while the
other two arcs for deletion are chosen so that the most favorable
quad change is obtained.
%-----------------------------------------------------------------
\section{Computational Experiments} \label{sec:experiments}
%-----------------------------------------------------------------
This section presents the results of
computational experiments
on the ATSP instances from TSPLIB library~\cite{Reinelt91}.
The GA was programmed in Java (NetBeans IDE~7.2.1) and tested on a
computer with Intel~Core~2~Duo~CPU~E7200 2.53~GHz processor, 2 Gb
RAM. In the experiments, we set the population size ${N=100}$, the
tournament size~${s=10}$ and the mutation
probability~${p_{mut}=0.1}$.

Our GA is restarted every time as soon as the current iteration
number becomes twice the  iteration number when the best
incumbent was found, until the overall execution time reached the
limit. Moreover, if the greedy heuristic of W.~Zhang generates
only one subcycle, this indicates that the ATSP instance was
solved to optimality, and the algorithm stops.  The best solution
found over all restarts was returned as the result. We have also
tested an alternative approach, where the GA runs for the whole
given period of time without restarts but it turned to be inferior
to the GA with the restart rule.

The first computational experiment is aimed at comparison of the performance of our
GA based on ODEC (${GA_{ODEC}}$) with SAX/RAI memetic algorithm (${MA_{SAX/RAI}}$)
from~\cite{BFM2004}, which has one of the best results in the literature on metaheuristics for the ATSP.
In order to put the considered algorithms into equal positions, ${GA_{ODEC}}$ was given
 the CPU-time~limit (denoted as $T$) by a factor~$3$ less than the CPU resource
used by ${MA_{SAX/RAI}}$ in~\cite{BFM2004}. This scaling factor
chosen on the basis a rough comparison of computers by means of performance table~\cite{Dongarra}.
For a statistical comparison, on each instance we executed
${GA_{ODEC}}$  $1000$ times. In each execution ${GA_{ODEC}}$ was
given the same CPU-time~limit indicated above. In~\cite{BFM2004},
${MA_{SAX/RAI}}$ was run $20$ times on each instance.
Table~\ref{tab:ATSP_GA+MA} shows the obtained results, where
$F_{\mathrm{opt}}$ represents the frequency of finding an optimum,
$\Delta_{\mathrm{err}}$~is the average percentage deviation of the
length of a resulting solution from the optimum,
$\Delta_{\mathrm{init}}$~denotes the average percentage deviation
of the length of the best initial solution from the optimum.
As seen from Table~\ref{tab:ATSP_GA+MA}, ${GA_{ODEC}}$ achieved
$100\%$ success rate on $17$ out of $26$ instances. On each
instance, ${GA_{ODEC}}$ found optima in not less than $91\%$~of runs.

The statistical analysis of experimental data was carried out
using a significance test of the null hypothesis
from~\cite{BH1977}, Ch.~8,~\S2. Suppose that two algorithms are
compared in terms of probability of ``success'', where ``success''
corresponds to finding an optimal solution. Let $P_1$ and $P_2$
denote the  probabilities of success for the considered
algorithms. The null hypothesis is expressed by $P_1=P_2$.

The test procedure is as follows. Under the null hypothesis, the
estimate of common  success rate is
$\hat{P}=\frac{\hat{P_1}N_1+\hat{P_2}N_2}{N_1+N_2}$, where
$\hat{P_1}$ denotes the frequency of success in $N_1$ runs for the
first algorithm and $\hat{P_2}$ is the frequency of success in
$N_2$ runs for the second algorithm. Then the difference
$\hat{P_1} - \hat{P_2}$ is expressed in units of the standard
deviation by calculating the statistic $A=\frac{|\hat{P_1} -
\hat{P_2}|}{\hat{SD}}$, where
$
\hat{SD}=\sqrt{\frac{\hat{P}(1-\hat{P})}{N_1}+\frac{\hat{P}(1-\hat{P})}{N_2}}
$
is the estimation of the standard deviation.

\begin{table}[!h]
\scriptsize
\caption{{Computational Results for the ATSP Instances}}%~
\label{tab:ATSP_GA+MA}
\begin{tabular}{|c|c|c|c||c|c|c|c||c||c|c|c|c|}
\hline
Instance & \multicolumn{8}{c||}{Genetic Algorithms} & \multicolumn{4}{c|}{${MA_{SAX/RAI}}$} \\
\cline{2-9}
 & \multicolumn{3}{c||}{${GA_{ODEC}}$} & \multicolumn{4}{c||}{${GA_{ER}}$} & $T$, & \multicolumn{4}{c|}{} \\
\cline{2-8}\cline{10-13}
&$\Delta_{\mathrm{init}}$ &  $F_{\mathrm{opt}}$ & $\Delta_{\mathrm{err}}$ &$\Delta_{\mathrm{init}}$ &  $F_{\mathrm{opt}}$ & $\Delta_{\mathrm{err}}$ & $A$& sec. & $\Delta_{\mathrm{init}}$ &  $F_{\mathrm{opt}}$ & $\Delta_{\mathrm{err}}$& $A$\\
\hline ftv33 & 0.00 & {\bf 1} & 0.00  & 0.00 & {\bf 1} & 0.00 & 0 & 0.097 &12.83& {\bf 1} & 0.00& 0\\
\hline ftv35 & 0.00 & {\bf 1} & 0.00 & 0.00 & {\bf 1} & 0.00 & 0 & 0.11 &0.14 & {\bf 1} & 0.00& 0\\
\hline ftv38 & 0.00 & {\bf 1$^*$} & 0.00  & 0.131 & {\bf 1} & 0.00 & 0 & 0.103 &0.13 & 0.25 & 0.10& 27.6\\
\hline p43   & 0.00 & {\bf 1$^*$} & 0.00 &  0.00 & {\bf 1} & 0.00 & 0  & 0.16 &0.05 & 0.55 & 0.01& 21.3\\
\hline ftv44 & 0.098 & {\bf 1$^*$} & 0.00  & 0.167 & 0.874 & 0.078 & 11.6 & 0.137 &7.01 & 0.35 & 0.44& 25.7\\
\hline ftv47 & 0.199 & {\bf 1} &0.00 & 0.338 & {\bf 1} & 0.00 & 0 &0.157 &2.70 & {\bf 1} & 0.00& 0\\
\hline ry48p & 0.978 & {\bf 0.997$^*$} & 0.0001 & 3.511 & 0.520 & 0.092 & 24.9 &0.187 & 5.42 & 0.85 & 0.03& 8.8\\
\hline ft53 & 0.438 & {\bf 1} & 0.00  & 5.073 & 0.668 & 0.035 & 19.9 &0.187 &18.20 & {\bf 1} & 0.00& 0\\
\hline ftv55 & 0.002 & {\bf 1} & 0.00  & 0.002 & {\bf 1} & 0.00 & 0 &0.167 & 3.61 & {\bf 1} & 0.00& 0\\
\hline ftv64 & 0.032 & {\bf 1} & 0.00 & 0.376 & 0.989 & 0.002 & 3.3 & 0.22 &3.81 & {\bf 1} & 0.00& 0\\
\hline ft70 &0.367 & {\bf 1$^*$} &0.00  & 0.321 & 0.583 & 0.013 & 22.9 &0.32 &1.88 & 0.4 & 0.03& 24.6\\

\hline ftv70 & 1.025 & {\bf 1$^*$} & 0.00  & 1.525 & 0.660 & 0.098 &  20.2 & 0.277 &3.33 & 0.95 & 0.01& 7.1\\

\hline ftv90 & 0.063 &  0.976 & 0.003  & 0.318 & 0.516& 0.007 & 23.6 &0.317 &3.67 & {\bf 1} & 0.00& 0.7\\
\hline ftv100 & 0.386 & 0.92 & 0.013 & 1.092 & 0.784 & 0.016 &  8.6 & 0.4 & 3.24 & {\bf 1} & 0.00& 1.3\\
\hline kro124p & 0.164 & {\bf 0.996$^*$} & 0.0001 & 0.288 & 0.322 & 0.033 & 31.8 & 0.457 &6.46 & 0.90 & 0.01& 5.6\\
\hline ftv110 & 0.287 & {\bf 0.972} & 0.003  & 0.305 & 0.854 & 0.025 & 9.4 &0.57 &4.7  & 0.90 & 0.02& 1.9\\
\hline ftv120 & 0.156 &  {\bf 0.912$^*$} & 0.008  & 2.463 & 0.430 & 0.506 & 22.9 &0.73 &8.31  & 0.35 & 0.14& 8.3\\
\hline ftv130 & 0.342 & {\bf 0.934} & 0.008  & 1.841 & 0.361 & 0.068 & 26.8 &0.727 &3.12  & 0.90 & 0.01& 0.6\\
\hline ftv140 & 0.111 & {\bf 0.947$^*$} & 0.004 & 0.601 & 0.463 & 0.065 & 23.7 & 0.887 &2.23  & 0.70 & 0.08& 4.7\\
\hline ftv150 & 0.739 & {\bf 0.982$^*$} & 0.002 & 1.358 & 0.532 & 0.068 & 23.5 & 0.897&2.3  & 0.90 & 0.01& 2.6\\
\hline ftv160 & 0.026 & {\bf 1$^*$} & 0.00  & 0.958 & 0.491 & 0.099 & 26.1 &1.093 &1.71  & 0.80 & 0.02& 14.2\\
\hline ftv170 & 0.108 & {\bf 1$^*$} & 0.00 & 0.334 & 0.222 & 0.141 & 35.7 & 1.307& 1.38  & 0.75 & 0.05& 15.9\\
\hline rbg323 & 0.00  & {\bf 1} & 0.00 & 0.00  & {\bf 1} & 0.00& 0 &0.03& 0.00  & {\bf 1} & 0.00& 0\\
\hline rbg358 & 0.00  & {\bf 1} & 0.00  & 0.00  & {\bf 1} & 0.00& 0 &0.03& 0.00  & {\bf 1} & 0.00&0\\
\hline rbg403 & 0.00  & {\bf 1} & 0.00  & 0.00  & {\bf 1} & 0.00& 0 &0.032& 0.00  & {\bf 1} & 0.00&0\\
\hline rbg443 & 0.00  & {\bf 1} & 0.00  & 0.00  & {\bf 1} & 0.00& 0 &0.033& 0.00  & {\bf 1} & 0.00&0\\
\hline Average & 0.212 & 0.986 & 0.0016 & 0.808 & 0.741 & 0.0518 & 12.9 &0.371&3.701 & 0.829 & 0.0369 & 6.6\\
\hline
\end{tabular}
\end{table}

It is supposed that statistic $A$ is normally distributed.
 To test the null hypothesis  versus the alternative one at
a confidence level~$\alpha$, we compare the computed $A$ to the
quantile  of standard normal distribution $z_{\alpha/2}$. If $A$
is lager than $z_{\alpha/2},$ the null hypothesis is rejected.
Otherwise the null hypothesis is accepted. At $\alpha=0.05$ we
have $z_{0.025}=1.96$. The values of statistic $A$ for algorithms
${GA_{ODEC}}$ and ${MA_{SAX/RAI}}$  are found and presented in
the last column of Table~\ref{tab:ATSP_GA+MA}
(`$*$' indicates the statistical significance difference between  ${GA_{ODEC}}$ and ${MA_{SAX/RAI}}$
 at level $\alpha=0.05$).

In $14$ out of $26$  instances, ${GA_{ODEC}}$ finds an optimum
more frequently than ${MA_{SAX/RAI}}$ (in $12$  cases among these,
the difference between the frequencies of finding an optimum is
statistically significant). Both algorithms demonstrate $100\%$
frequency of obtaining an optimum on $10$ problems. Note that the
heuristic of W.~Zhang is very efficient on series rbg and the
optimal solutions to all rbg instances were found in the
considered algorithms at the initialization stage.
${MA_{SAX/RAI}}$ slightly outperforms ${GA_{ODEC}}$ only on two
instances ftv90 and ftv100, but the differences
are not statistically significant.
Moreover,  the average quality of the resulting solutions for
${GA_{ODEC}}$ is approximately in $23$ times better than the
average quality for~${MA_{SAX/RAI}}$. The  quality of initial
solutions is better in our algorithm. (Note that we use the local
search at the initialization stage, while ${MA_{SAX/RAI}}$ applies
a local search only on GA iterations.)

Recently, Tin\'{o}s at el.~\cite{TWO14} proposed a GA with new
crossover operator~GAPX, which presents very competitive results
in terms of frequencies of finding an optimum, but its CPU
resource usage  is significantly higher than that of
${GA_{ODEC}}$.  On all of $16$ TSPLIB-instances tested in~\cite{TWO14} GA with GAPX demonstrated $100\%$ success,
while ${GA_{ODEC}}$ displayed $99.96\%$ success on average.
However, the average CPU-time~$T$ of our GA was $0.22$~sec.  on these instances, and
the overall CPU-time of GA with GAPX was $98.38$~sec. on a similar computer.

In the second experiment, we compare our steady state GA to the
similar GA  with the population management strategy known as {\em
elitist recombination}~\cite{GoldTheir94} (${GA_{ER}}$) under the
same CPU-time limit. The results are also listed in
Table~\ref{tab:ATSP_GA+MA}. The eighth column represents the values
of statistic $A$ for comparison of ${GA_{ER}}$ against
${GA_{ODEC}}$ on all ATSP instances. We estimate the average
frequency of finding optimal solutions for ${GA_{ER}}$ as
approximately $60\%$ of the average frequency for ${GA_{ODEC}}$
(the difference between the frequencies is statistically
significant), except for $10$ of $26$ instances where both
algorithms have $100\%$  success. Note that the GA with elitist
recombination maintains the population diversity better. Due to
this reason, the GA with elitist recombination outperformed the
steady state GA in our preliminary experiments with no restarts,
which were organized analogously to the experiments
in~\cite{EK2014II}. The restarts performed in ${GA_{ODEC}}$ allow
to avoid localization of the search and restore the population
diversity, leading to better results.

We carried out the third experiment in order to compare the
optimized crossover ODEC to its randomized prototype DEC. This
experiment clearly showed an advantage of ODEC over DEC.  The
modification of ${GA_{ODEC}},$ where operator DEC substitutes
ODEC, on average gave only  $45\%$ frequency of obtaining an
optimum within the same CPU~time limit. Moreover, for the
large-scale problems such as ftv120,
ftv130, ftv140, ftv150 and ftv170 the  GA with DEC found optimal solution  no more than once out of 1000~runs.

We also estimate that the average frequency of success of the GAs
with optimal recombination reported in~\cite{EK2014II} is twice as
small compared to such frequency for ${GA_{ODEC}}$, even though
the GAs in~\cite{EK2014II} were given more CPU~time.

%-----------------------------------------------------------------
\section{Conclusions}\label{sec:concl}
%-----------------------------------------------------------------

We proposed a steady-state GA with adjacency-based representation
using an optimal recombination and a local search to solve the
ATSP. An experimental evaluation on instances from TSPLIB library
shows that the proposed GA  yields results competitive to those of
some other state-of-the-art genetic algorithms. The experiments also indicate % GAs
that the proposed GA dominates a similar GA based on the
population management strategy, known as elitist recombination.
The restarts performed in the proposed GA allow to avoid
localization of search and restore the population diversity,
leading to better results when the steady-state population
management is used. The experiments also show an advantage of the
deterministic optimized crossover over its randomized prototype.

%\section*{Acknowledgements}
%This research is supported by the Russian Science
%Foundation grant 15-11-10009.

\renewcommand{\refname}{References}
\providecommand{\url}[1]{\texttt{#1}}
\providecommand{\urlprefix}{URL }


\begin{thebibliography}{10}
\providecommand{\url}[1]{\texttt{#1}}
\providecommand{\urlprefix}{URL }

\bibitem{BH1977}
Brown, B.W., Hollander, M.: Statistics: A Biomedical Introduction. John Wiley
  \& Sons, Inc (1977)

\bibitem{BFM2004}
Buriol, L.S., Franca, P.M., Moscato, P.: A new memetic algorithm for the
  asymmetric traveling salesman problem. Journal of Heuristics  10,  483--506
  (2004)

\bibitem{CS03}
Cook, W., Seymour, P.: Tour merging via branch-decomposition. INFORMS Journal
  on Computing  15(2),  233--248 (2003)

\bibitem{Dongarra}
Dongarra, J.J.: Performance of various computers using standard linear
  equations software. Tech. Rep. CS-89-85, University of Manchester (2014), 110
  p.

\bibitem{Epp2007}
Eppstein, D.: The traveling salesman problem for cubic graphs. Journal of Graph
  Algorithms and Applications  11(1) (2007)

\bibitem{EK2014IIY}
Eremeev, A.V., Kovalenko, J.V.: Optimal recombination in genetic algorithms for
  combinatorial optimization problems: Part $\mbox{II}$. Yugoslav Journal of
  Operations Research  24(2),  165--186 (2014)

\bibitem{EK2014II}
Eremeev, A.V., Kovalenko, J.V.: Experimental evaluation of two approaches to
  optimal recombination for permutation problems. In: 16th European Conference,
  EvoCOP 2016, LNCS. pp. 138--153. Porto, Portugal (2016)

\bibitem{GJ}
Garey, M.R., Johnson, D.S.: Computers and Intractability. A Guide to the Theory
  of NP-completeness. W.~H.~Freeman and Company, San Francisco (1979)

\bibitem{GoldTheir94}
Goldberg, D., Thierens, D.: Elitist recombination: An integrated selection
  recombination $\mbox{GA}$. In: First IEEE World Congress on Computational
  Intelligence,. vol.~1, pp. 508--512. IEEE Service Center, Piscataway, New
  Jersey (1994)

\bibitem{Johnson97}
Johnson, D.S., McGeorch, L.A.: The traveling salesman problem: a case study.
  In: Aarts, E., Lenstra, J.K. (eds.) Local Search in Combinatorial
  Optimization, pp. 215--336. John Wiley \& Sons Ltd. (1997)

\bibitem{KP1980}
Kanellakis, P.C., Papadimitriou, C.H.: Local search for the asymmetric
  traveling salesman problem. Oper. Res.  28,  1086--1099 (1980)

\bibitem{Karp1979}
Karp, R.M.: A patching algorithm for the nonsymmetric traveling-salesman
  problem. SIAM Journal on Computing  8,  561--573 (1979)

\bibitem{R94}
Radcliffe, N.J.: The algebra of genetic algorithms. Annals of Mathematics and
  Artificial Intelligence  10(4),  339--384 (1994)

\bibitem{Reev1997}
Reeves, C.R.: Genetic algorithms for the operations researcher. INFORMS Journal
  on Computing  9(3),  231--250 (1997)

\bibitem{Reinelt91}
Reinelt, G.: $\mbox{TSPLIB}$ -- a traveling salesman problem library. ORSA
  Journal on Computing  3(4),  376--384 (1991)

\bibitem{TWO14}
Tin\'{o}s, R., Whitley, D., Ochoa, G.: Generalized asymmetric partition
  crossover ($\mbox{GAPX}$) for the asymmetric $\mbox{TSP}$. In: The 2014
  Annual Conference on Genetic and Evolutionary Computation. pp. 501--508. ACM
  New York, NY (2014)

\bibitem{S91}
Whitley, D., Starkweather, T., Shaner, D.: The traveling salesman and sequence
  scheduling: Quality solutions using genetic edge recombination. In: Davis, L.
  (ed.) Handbook of Genetic Algorithms, pp. 350--372. Van Nostrand Reinhold, NY
  (1991)

\bibitem{YI}
Yagiura, M., Ibaraki, T.: The use of dynamic programming in genetic algorithms
  for permutation problems. Eur. Jour. Oper. Res.  92,  387--401 (1996)

\bibitem{Zhang00}
Zhang, W.: Depth-first branch-and-bound versus local search: A case study. In:
  17th National Conf. on Artificial Intelligence. pp. 930--–935. Austin, TX
  (2000)

\end{thebibliography}
\end{document}